\documentclass{article}

\usepackage{arxiv}
\usepackage[utf8]{inputenc} 
\usepackage[T1]{fontenc}  
\usepackage{hyperref}      
\usepackage{url}           
\usepackage{booktabs}       
\usepackage{amsfonts}       
\usepackage{nicefrac}       
\usepackage{microtype}     
\usepackage{lipsum}
\usepackage{times}
\usepackage{epsfig}
\usepackage{graphicx}
\usepackage{amsmath}
\usepackage{amssymb}
\usepackage{diagbox}
\usepackage{subcaption}
\usepackage{lineno}
\usepackage{biblatex}
\usepackage{algorithm} 
\usepackage{algpseudocode} 
\usepackage{amssymb}
\usepackage{caption} 
\usepackage[flushleft]{threeparttable}
\usepackage{setspace}
\usepackage{wrapfig}
\usepackage[dvipsnames]{xcolor}
\usepackage{bm}

\title{On the link between generative semi-supervised learning and generative open-set recognition}

\author{
  Emile-Reyn Engelbrecht \\
  Main and corresponding author \\
  Electronic Engineering \\
  Stellenbosch University \\
  South Africa \\
  \texttt{18174310@sun.ac.za} \\
   \And
  Johan A. du Preez \\
  Co-author \\
  Electronic Engineering \\
  Stellenbosch University \\
  South Africa \\
  \texttt{dupreez@sun.ac.za} \\
}

\begin{document}
\maketitle
\begin{abstract}
This study investigates the relationship between semi-supervised learning (SSL, which is training off partially labelled datasets) and open-set recognition (OSR, which is classification with simultaneous novelty detection) under the context of generative adversarial networks (GANs). Although no previous study has formally linked SSL and OSR, their respective methods share striking similarities. Specifically, SSL-GANs and OSR-GANs require their generators to produce 'bad-looking' samples which are used to regularise their classifier networks. We hypothesise that the definitions of bad-looking samples in SSL and OSR represents the same concept and realises the same goal. More formally, bad-looking samples lie in the complementary space, which is the area between and around the boundaries of the labelled categories within the classifier's embedding space. By regularising a classifier with samples in the complementary space, classifiers achieve improved generalisation for SSL and also generalise the open space for OSR. To test this hypothesis, we compare a foundational SSL-GAN with the state-of-the-art OSR-GAN under the same SSL-OSR experimental conditions. Our results find that SSL-GANs achieve near identical results to OSR-GANs, proving the SSL-OSR link. Subsequently, to further this new research path, we compare several SSL-GANs various SSL-OSR setups which this first benchmark results. A combined framework of SSL-OSR certainly improves the practicality and cost-efficiency of classifier training, and so further theoretical and application studies are also discussed.
\end{abstract}

% keywords can be removed

%labelling vs annotation
%Classifier vs model

\section{Introduction}
Classifier networks are trained to categorise input data samples into pre-labelled categories. However, classifiers must also be equipped to detect unknown or novel categories that may emerge over time~\cite{geng2020recent}. Two critical applications of novelty detection are 1) automated diagnostic tools~\cite{pahar2021automatic} and 2) self-driving cars~\cite{wu2019semi}. In both cases, misclassifications can lead to fatal consequences. Therefore, to ensure safe classifications, classifiers must be able to detect and separate novel categories that were not present during training but appeared during testing. As per the examples, 1) classifiers should call on human doctors when encountering an unknown disease (e.g. SARS-CoV-2 pre-December 2019~\cite{andersen2020proximal}), and 2) classifiers must alert driving modules to safely manoeuvre out of the way or demand manual takeover in case of unexpected driving scenes (e.g. unexpected obstacles~\cite{ramos2017detecting}, or high-risk scenarios~\cite{puertas2021should}).

Open-set recognition (OSR) considers classification with simultaneous novelty detection. More specifically, OSR tests classifiers' ability to handle both labelled and novel categories unobserved during training~\cite{geng2020recent}. To do so, OSR defines an open testing domain, meaning trained classifiers must accurately categorise test samples belonging to one of $K$ labelled categories and separate test samples belonging to unobserved novel categories into an additional $K + 1$'th unknown category. It is important to note that OSR is independent of the training method used since it only inserts novel categories during the testing phase. However, OSR studies generally apply supervised learning to train classifiers, which is known to be costly in many real-world applications~\cite{van2020survey}. Instead, training under semi-supervised learning (SSL, which is training off partially labelled datasets) and testing classifiers under the OSR regime would significantly improve the cost-efficiency and practicality of classifier development. 

\begin{figure*}[!t]
\centering
\subfloat[Example domain]{\includegraphics[width=0.35\textwidth]{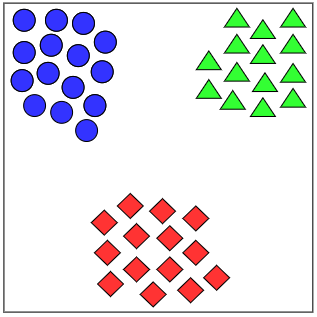}
\label{fig:1a}}
\hfil
\centering
\subfloat[Classification boundaries]{\includegraphics[width=0.35\textwidth]{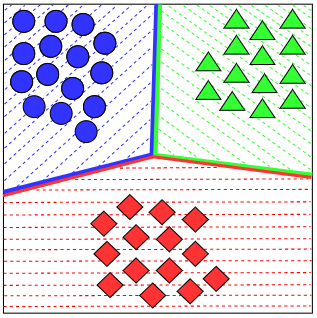}
\label{fig:1b}}
\caption{Different classification boundaries of a 2D example domain with $K = 3$ number of labelled categories: (a) visually describes the 2D domain with each category represented by a different shape and colour; and (b) describes the classification boundaries of a typical closed set classifier with a softmax output activation function.}
\label{fig:1}
\end{figure*}

\begin{figure*}[!t]
\centering
\centering
\subfloat[Complementary space (SSL)]{\includegraphics[width=0.35\textwidth]{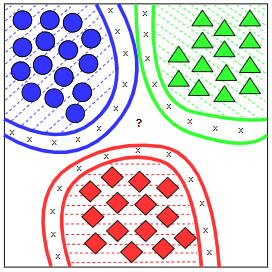}
\label{fig:2a}}
\hfill 
\centering
\subfloat[Open space (OSR)]{\includegraphics[width=0.35\textwidth]{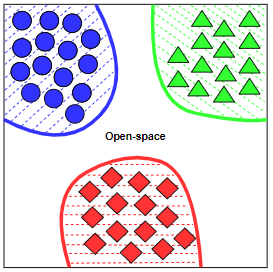}
\label{fig:2b}}
\caption{Visual representation of the embedding spaces generalised within different fields: (a) shows the complementary space of each category as required for semi-supervised learning (SSL), which specifically depicted by the 'x's around each category; and (b) shows the open space of the domain as required for open-set recognition (OSR), which is represented by the white area.}
\label{fig:2}
\end{figure*}

SSL and OSR are rarely linked within research. However, SSL and OSR share striking similarities in their proposed training methods, particularly when using generative adversarial networks (GANs) (SSL -~\cite{NIPS2016_8a3363ab, dai2017good, NEURIPS2019_517f24c0, li2020semi} and OSR -~\cite{neal2018open, jo2018open, chen2021adversarial}). Consider a typical closed-set classifying network with a softmax output activation function for $K = 3$ categories depicted in Fig.~\ref{fig:1a} and Fig.~\ref{fig:1b}. A conventional closed-set classifier learns an embedding space that covers the entire domain. In other words, the model embeds different classification boundaries for each category that would maximally separate the categories from one another, leaving no space open. However, maximising boundaries is insufficient for SSL~\cite{dai2017good} and OSR~\cite{chen2021adversarial}. More specifically, SSL classifiers must also generalise the complementary space~\cite{NIPS2016_8a3363ab, dai2017good} (see Fig. \ref{fig:2a}), and OSR classifiers must generalise the open space~\cite{chen2021adversarial} (see Fig. \ref{fig:2b}). This study will show that generalising the complementary space inherently generalises the open space as well. As to say, SSL-GANs and OSR-GANs achieve the same goal. 

The specific methodology behind SSL-GANs and OSR-GANs is to generalise an additional $K + 1$'th category to represent the necessary embedding spaces as theorised in each field. As previously mentioned, SSL-GANs aim to generalise the complementary space in the $K + 1$'th category and OSR-GANs aim to generalise the open space in the $K + 1$'th category. To do so, SSL-GANs and OSR-GANs design their generators to produce samples that, when pseudo-labelled into the $K + 1$'th category, ensures that the classifier will generalise the necessary embedding spaces. However, it is interesting that SSL-GANs and OSR-GANs rely on these generated samples looking 'bad' by some theoretical definition. This study postulates that bad-looking generated samples in both SSL-GAN and OSR-GAN studies are the same in that both lie in the complementary space. Then, when pseudo-labelling generated samples in the complementary space into the $K + 1$'th category, SSL and OSR classifiers both generalise the open space. 
 
To prove that SSL-GANs and OSR-GANs operate in the same manner, two GAN models are theoretically described and experimentally compared. For SSL, we study feature-matching (FM)-GANs~\cite{NIPS2016_8a3363ab}, which set the foundation of complementary space theory. For OSR, we study adversarial reciprocal point (ARP)-GANs~\cite{chen2021adversarial}, which exhausts open space theory to achieve state-of-the-art OSR results. Each GAN is thoroughly described concerning its setup and loss functions, specifically emphasising their definitions of the $K + 1$'th category. These descriptions make it clear that both models aim to produce samples in the complementary space, and that both models define their $K + 1$'th categories as a regularisation category. In other words, SSL-GANs and OSR-GANs use bad-looking generated samples to regularise their classifying networks. To prove these findings, FM-GANs and ARP-GANs are compared under the same experimental setups. Considering that both achieve near identical results, we can conclude the SSL-OSR link under the GAN context. 

Given the SSL-OSR link, it is also important to develop benchmark results for this newly discovered field. Consequently, we experiment with several GANs to determine the best performer for SSL-OSR experiments. Specifically, we experiment with Bad-GANs~\cite{dai2017good}, Margin-GANs~\cite{NEURIPS2019_517f24c0} and Triple-GANs~\cite{li2021triple}, and find that Margin-GANs outperform the other models. Finally, we also discuss various applications of SSL-OSR, with the aim of promoting this field to a variety of research spaces. The remainder of this paper is structured as follows - Section~2 provides background information on SSL, OSR, and the previous research on the links between these fields; Section~3 provides the necessary theory on FM-GANs and ARP-GANs; Section~4 conducts experiments; Section~5 discusses applications of the new SSL-OSR field; and Section~6 concludes the study. 

%+- 1000 words

\section{Background}
Open-set recognition (OSR) is the research field that studies classification with simultaneous novelty detection. More specifically, OSR classifiers must accurately categorise samples that belong to $K > 1$ number of labelled categories and simultaneously separate samples that belong to any number of novel categories into an additional $K + 1$'th category. Novel categories are formally defined as groups of anomalous samples with similar patterns that do not match any of the labelled categories~\cite{gruhl2021novelty}. Within OSR, novel categories are further described as those that are unobserved during training but appear during testing~\cite{pimentel2014review, geng2020recent}. In other words, OSR ensures classifiers can handle changing environments wherein new patterns emerge over time~\cite{din2021data}. With regards to the training policy, OSR studies generally use supervised learning wherein the training set, $\mathcal{D}_{\text{lab-train}}$, contains sample pairs, $(x, y)$, where $x$ is the input sample and $y$ is the output category label. After training, an independent testing set, $\mathcal{D}_{\text{test}}$, is used to determine the performance of the classifier, wherein the novel categories could lie.

Testing sets also consist of sample pairs, $(x, y_a)$. However, considering that the classifier cannot access the testing labels, these are defined as anticipated labels, $y_a$, which are used to determine classifier performance. It is important to state that the testing sets should mimic real-world scenarios as closely as possible to ensure credible performance measurements. Let the set $C_K := \{1, 2, ..., K\}$ represent all labelled categories, with all entries in $C_K$ representing the category labels. Within conventional supervised learning, all training and testing samples belong to one of the categories in $C_K$. In other words, all $(y \sim \mathcal{D}_{\text{lab-train}}) \in C_K$ and all $(y_a \sim \mathcal{D}_{\text{test}}) \in C_K$. To extend supervised learning to OSR, samples from novel categories are inserted into the testing set. Let the $K + 1$'th category be defined as the augmented unknown category, meaning it is an encapsulating category representing any and all unobserved novel categories. Then, the OSR regime would have all $(y \sim \mathcal{D}_{\text{lab-train}}) \in C_K$ and all $(y_a \sim \mathcal{D}_{\text{test}}) \in (C_K \cup \{K + 1\})$. In other words, beyond generalising the $K$ labelled categories, OSR classifiers must generalise the $K + 1$'th augmented unknown category, noting that this category does not have any training samples. 

Supervised learning requires many labelled training samples to train classifiers, making it costly in real-world applications. Semi-supervised learning (SSL) proposes a more cost-efficient solution by training classifiers using a small labelled and a large unlabelled training set, $|\mathcal{D}_{\text{lab-train}}| << |\mathcal{D}_{\text{unlab-train}}|$~\cite{van2020survey}. Similar to the test set, the classifier cannot access the labels in $\mathcal{D}_{\text{unlab-train}}$, and so these labels are also represented as anticipated labels, $y_a$. We also focus on the inductive SSL setting, where the testing set is again independent of the training set. Conventional SSL relies on the cluster assumption, meaning we assume that all unlabelled training samples belong to one of the labelled categories - i.e. all $(y_a \sim \mathcal{D}_{\text{unlab-train}}) \in C_K$. Furthermore, conventional SSL assumes that all $(y_a \sim \mathcal{D}_{\text{test}}) \in C_K$, meaning there are no novel categories included in this learning policy. Similar to supervised learning, we can extend SSL to OSR by simply including novel categories in the test set - i.e. all $(y_a \sim \mathcal{D}_{\text{test}}) \in (C_K \cup \{K + 1\})$. This so-called SSL-OSR setting is the primary focus of this study. However, it is essential to note that as soon as we enter the realm of SSL, novel categories could also exist in $\mathcal{D}_{\text{unlab-train}}$. 

SSL-OSR assumes that novel categories only appear during testing. However, so-called observed novel categories have become more prevalent within SSL based research~\cite{engelbrecht2020open, yu2020multi}. Specifically, given the immense cost to label every category big data domains, recent studies have focused on the setting where observed novel categories are scattered in $\mathcal{D}_{\text{unlab-train}}$. This research field is called learning with augmented category by exploiting unlabelled data (LACU)~\cite{engelbrecht2020learning}. In this policy, the $K + 1$'th category is defined as a background augmented category to encapsulate all observed novel categories. It is important to note that LACU and OSR have contrasting definitions for the $K + 1$'th category since LACU considers observed novel categories and OSR considers unobserved ones. Although LACU represents an important SSL setting, this study only focuses on linking conventional SSL and OSR, meaning we assume that there are no novel categories seen in $\mathcal{D}_{\text{unlab-train}}$. Furthermore, our $K + 1$'th category is the augmented unknown category, as per the definition of OSR. 

GANs have been extensively used for conventional SSL~\cite{sajun2022survey}. Initially, SSL-GANs transformed the discriminator to distinguish between real and fake samples and simultaneously classify any real samples into its corresponding labelled category~\cite{odena2016semi, NIPS2016_8a3363ab, dai2017good, xu2021semi}. However, state-of-the-art models now prefer a three-network game, wherein the discriminator remains unchanged to the original GAN and a separate classifier is added to the mix~\cite{NEURIPS2019_517f24c0, li2021triple}. Of all SSL-GANs, the bad-GAN by Dai et al.~\cite{dai2017good} is especially noteworthy, as Dai theoretically and experimentally showed that good SSL-GANs should generate 'bad-looking' samples that lie in the complementary space. More specifically, Dai showed that regulating an SSL classifier using generated samples that lie in the complementary space (see Fig.~\ref{fig:2a}), then the classifier is forced to tighten the boundaries in the embedding space of labelled categories for improved generalisation. Bad-GANs are particularly interesting since OSR studies have used the same notion of bad-looking samples~\cite{sung2019difference}.

Before exploring OSR-GANs, it is essential to consider the use of GANs for conventional novelty detection - i.e. without multi-category classification. Conventional novelty detection is a one-class classification problem that distinguishes between unobserved novel categories from a single positive labelled category that represents the entire training domain~\cite{pimentel2014review}. The discriminator networks in GANs have been widely studied in this context~\cite{kliger2018novelty, perera2019ocgan, zhang2021adversarially}. More specifically, because the discriminator is tasked to separate fake-generated samples from all training samples, the discriminator can be considered a novelty detector if fake-generated samples represent in the open domain. However, the equilibrium theory of GANs suggests that a trained discriminator won't be able to determine whether an input sample is real or fake~\cite{NIPS2014_5ca3e9b1}. Therefore, unique adjustments are required to use a discriminator as a novelty detector. This study focuses on OSR, which extends novelty detection to also classify multiple labelled categories. Nevertheless, the underlying theory between OSR and novelty detection remains the same. 

A standard method to ensure that the discriminator/classifier can determine whether a sample is from the training domain or not is to design the generator to produce bad-looking samples~\cite{perera2019ocgan, sabokrou2018adversarially, neal2018open, jo2018open, chen2021adversarial}. Although multiple definitions have been proposed for these bad-looking samples, we argue that all definitions really rely on the SSL theory of the complementary space. Examples of such OSR-GANs include the counterfactual (counter)-GAN by Neal et al.~\cite{neal2018open}, which combines an auto-encoder and a GAN to generate counterfactual (or bad-looking) images that belong to one category but is pulled towards another. In contrast, Jo et al.~\cite{jo2018open} use a feature-matching (FM)-GAN and a denoising auto-encoder to generate samples that are corrupted with the noisy distribution of the feature space of the classifier. Finally, the recent adversarial reciprocal point learning GAN (ARP-GAN) proposed by Chen et al. also uses bad-looking samples that match reciprocal points~\cite{chen2021adversarial}. We note that ARP-GANs significantly outperformed all other OSR-GANs, and will be discussed in detail in the upcoming section. 

Although SSL-GANs and OSR-GANs theorise better generalisation and results given bad-looking generated samples, these methods have not yet been compared within research. Our hypothesis states that the notion of bad-looking samples in either the SSL or the OSR context represents samples that fall within the complementary space. In other words, counterfactual samples~\cite{neal2018open}, samples modelled with noisy distributions~\cite{jo2018open}, and samples based on reciprocal points~\cite{chen2021adversarial}, all allow classifiers to generalise the open space for the same reasons SSL models perform better when generalising the complementary space. That is, regularising a classifying network with samples in the complementary space inherently generalises the open space within the classifying network. With this hypothesis, an SSL-GAN would be capable at OSR, and an OSR-GAN would also be capable at SSL. Furthermore, all these models could be used for the combined task of SSL-OSR. 

Finally, it is important to note several studies that also focused on the combined task of SSL-OSR. Capazzo et al.~\cite{cappozzo2020anomaly} used traditional machine-learning techniques to address label noise under the SSL-OSR setting. However, these traditional techniques are unsuitable for high-dimensional data, and so these methods cannot be used for image classification. Another attempt at SSL-OSR is in the unpublished study by Kliger et al.~\cite{kliger2018novelty2}, who also applied GANs in this context. However, their proposed method and results were unclear, and no theoretical link was made between SSL-GANs and OSR-GANs. Finally, the most similar study to ours is the study by Sung et al.~\cite{sung2019difference}. Sung applied manually chosen operations to the training data to create bad-looking samples. Subsequently, FM-GANs were used to generate samples that matched fell between pseudo-data and the real-data which were used to regularise either a SSL or OSR classifying network. However, Sung did not justify the choice of operations to create the pseudo-data (i.e. no theoretical framework was developed for bad-looking samples), while Sung experimented on SSL and OSR independently without linking the two. In this study, we theoretically and experimentally link SSL with OSR under the unified framework of highly parameterised GANs.

%+- 1300 words

\section{GAN comparisons}
This section describes FM-GANs (SSL) and ARP-GANs (OSR) while explicitly highlighting their theoretical similarities. It is essential to pay attention to the definitions of the $K + 1$'th categories which both models use to represent unique areas in their classifier's embedding space. Specifically, FM-GANs use the $K + 1$'th category to represent the complementary space (as shown in Fig. \ref{fig:2a}), and ARP-GANs use the $K + 1$'th category to represent the open space (as shown in Fig. \ref{fig:2b}). However, both studies define the $K + 1$'th category as a so-called regularisation category instead of an additional label. Therefore, we propose that the $K + 1$'th categories in these different models ultimately represent the same theory, meaning the classifiers in FM-GANs and ARP-GANs function similarly. Please note that an understanding of the original GAN models is required for this discussion~\cite{NIPS2014_5ca3e9b1}. 

\subsection{Feature-matching GANs (SSL)}
Feature-matching (FM)-GANs provide vital insights into the link between SSL and OSR. FM-GANs were originally designed for SSL~\cite{NIPS2016_8a3363ab}, noting multiple follow-up studies~\cite{dai2017good, sung2019difference, li2020semi}. However, these models are interesting since they have also been used in multiple OSR studies~\cite{jo2018open, sung2019difference}. In this section, we describe FM-GANs as initially designed for SSL. However, we will explicitly focus on the theory of 'bad-looking' samples for improved SSL performance~\cite{dai2017good}. FM-GANs are one of the the original SSL-GANs that transforms the discriminator of the original GAN framework into a classifier network. In other words, the discriminator/classifier must distinguish between real or fake samples and classify all real samples into one of the $K$ labelled categories. However, extending the discriminator to also act as a classifier requires unique loss functions.  

The classifier network within FM-GANs, denoted as $C(x)$ for input sample $x$, is commonly described as having $K + 1$ output nodes. $K$ nodes are used for the labelled categories, while the $K + 1$'th node is used to detect fake samples. Following conventional supervised neural networks, the output of such a $K + 1$ classifier for a sample belonging to class $j$ would be given as \mbox{$p_{K + 1}(y = j | x) = \frac{\text{exp}[C^{j}(x)]}{\sum_{i=1}^{K + 1} \text{exp}[C^{i}(x)]}$}, where \mbox{$C^{i}(x)$} is the logit value of node $i$ and \mbox{$j \in (C_K \cup \{K + 1\})$}. Given this probability function, the FM-GAN's classifier minimises a combined supervised loss function for labelled training samples, and a supervised loss for fake-generated samples which are pseudo-labelled into the $K + 1$'th category. Specifically, this loss is described as:

\begin{equation}
\begin{split}
 \underset{\text{SSL}}{\text{D/C-Loss}} = 
 & - \; \mathbb{E}_{z \sim \mathbb{P}^{\text{(z)}}}\big[ \log( \; p_{\text{class}}(y = K + 1| \; G(z))) \big] \\
 & - \; \mathbb{E}_{(x, y) \sim D_{\text{lab-train}}} \big[
 \log(p_{\text{class}}(y = y|x) \;) \big].
    \label{eq:cost_c_ssl}
\end{split}
\end{equation} 

For such a $K + 1$ classifier, the probability of a sample being fake (i.e. being classified into the $K + 1$'th category) is given as \mbox{$p_{K + 1}(y = K + 1 | x) = \frac{\text{exp}[C^{K + 1}(x)]}{\sum_{i=1}^{K+1} \text{exp}[C^{i}(x)]}$}. However, FM-GANs found a different setup more conducive for SSL. Specifically, instead of defining the classifier to have an additional $K + 1$ output label, FM-GANs define the $K + 1$'th category as a so-called regularisation category. Consider again \mbox{$p_{K + 1}(y = j | x)$}, which is the softmax output activation function over $K + 1$ nodes. This function is over parameterised, meaning that subtracting any fixed vector from each output logit, $C^i$, does not change the probability scores. Consequently, we can fix any single output logit to the zero'th vector of the neural network's output. FM-GANs opt to fix the $K + 1$'th category to the zero'th vector, i.e. \mbox{$C^{K+1}(x) = 0 \; \forall x$}. In this case, the classifier only has $K$ ouput nodes, and the probability of a sample being fake changes to \mbox{$p_{\text{FM}}(y = K + 1 | x) = \frac{1}{\sum_{i=1}^{K} \text{exp}[C^{i}(x)] + 1}$}. 

When fixing the $K + 1$'th category to the zero'th vector, we can no longer use a supervised loss function for the generated samples. However, it is important to note that any input sample can only be real or fake, but never both. Therefore, the probability of the sample being real is given as \mbox{$p_{\text{FM}}(y \neq K + 1 | x) = (1 - p_{\text{FM}}(y = K + 1 | x)) = \frac{\sum_{i=1}^{K} \text{exp}[C^{i}(x)]}{\sum_{i=1}^{K} \text{exp}[C^{i}(x)] + 1}$}. As per the original GAN criteria (i.e. $K = 0$ or a one-class classifier), the discriminator would maximise \mbox{$\log(p_{\text{FM}}(y = K + 1 | x))$} for fake generated samples and maximise \mbox{$\log(p_{\text{FM}}(y \neq K + 1 | x))$} for real training samples. FM-GANs apply the same tactic, but under the multi-category setting. Again, we couple the real-fake criteria with a supervised loss for labelled training samples and get the following loss function for the FM-GAN's discriminator/classifier:

\begin{equation}
\begin{split}
 \underset{\text{FM-GAN}}{\text{D/C-Loss}} = 
  & - \mathbb{E}_{z \sim \mathbb{P}^{\text{(z)}}} \big[ \log \bigg(\frac{1}{\sum_{i=1}^{K} \text{exp}[C^{i}(G(z))] + 1} \bigg) \big]  \\ 
 & - \mathbb{E}_{x \sim D_{\text{unlab-train}}} \big[ \log \bigg(\frac{\sum_{i=1}^{K} \text{exp}[C^{i}(x)]}{\sum_{i=1}^{K} \text{exp}[C^{i}(x)] + 1} \bigg) \big]\\
 & - \; \mathbb{E}_{(x, y) \sim D_{\text{lab-train}}} \big[ \log(p_{\text{class}}(y|x) \;) \big] .
    \label{eq:cost_c_fm1}
\end{split}
\end{equation}

By minimising the above loss function, the discriminator/classifier in an FM-GAN is trained to 1) produce low output logits for fake-generated samples, 2) produce high output logits for real training samples, and 3) ensure that the largest output logit corresponds to the category label of the input sample. Regularisation is the process of reducing the confidence of a classifier's predictions~\cite{gouk2021regularisation}. By training the classifier to produce low output logits for fake-generated samples, we argue that FM-GANs are really regularising the classifier with the generated samples. Similar regularisation mechanisms have been used in other SSL-GANs, including the inverse-entropy in Margin-GANs~\cite{NEURIPS2019_517f24c0} and the pseudo discriminative loss in Triple-GANs~\cite{li2021triple}. It is also important to note that the generated samples are ever-evolving in the GAN training process. However, if we maintain the original GAN criteria, and the generated samples end up matching the probability distribution of the training data, then the regularisation in FM-GANs would hinder the generalisation of the classifier~\cite{dai2017good}.

To ensure the generator produces samples that would benefit the regularisation of the discriminator/classifier, FM-GANs task the generator to produce samples that lie in the complementary space (see Fig.~\ref{fig:2a}). Specifically, FM-GANs propose feature-matching for the generator network. The FM loss function is based on the proposition that the last hidden layer of the classifier network (denoted as $C'(x)$) represents the features of the input data. To match the features between the real training samples and the fake generated samples, the generator within an FM-GAN minimises the L2 distance between the activations of this hidden layer for real training samples and the activations for fake generated samples. Although the FM-loss function does not theoretically guarantee that generated samples will lie in the complementary space, it does ensure that generated samples will always look 'bad'. In other words, the generator can only learn a partial representation of the training distribution, ensuring that the classifier's regularisation will not affect the learning process. Formally, the FM-GAN generator loss function is given as follows:

\begin{equation}
\label{eq:cost_g_fm}
 \underset{\text{FM-GAN}}{\text{G-Loss}} \; = \;  \mathbb{E}_{x \sim D_{\text{train}} \; , \; z \sim \mathbb{P}^{\text{(z)}}} \big[ || C'(x) -  C'(G(z)) ||^2 \big].
\end{equation}

In summary, FM-GANs convert the traditional GAN game into a SSL framework by designing the generator to produce bad-looking samples which are used to regularise the discriminator/classifier. Although various improvements have been proposed to further push generated samples into the complementary space~\cite{dai2017good}, we focused on FM-GANs for several reasons: 1) FM-GANs have enhanced stability and so can be easily inserted into different code bases, and 2) FM-GANs have also been used in OSR studies~\cite{jo2018open}. Our hypothesis states that by designing the $K + 1$'th category as a regularisation category, FM-GANs will also be highly capable at novelty detection. As to say, lowering the confidence of the classifier network will provide enhanced classification with a reject option, as is standard in OSR. To further emphasise this SSL-OSR link, we next consider the current state-of-the-art OSR model. 

%Word count: 1100

\subsection{Adversarial reciprocal point GANs (OSR)}
OSR classifiers generalise the open space, which is 'all that the training distribution is not'. Several studies have proposed generative methods to generalise the open space within the additional $K + 1$'th category~\cite{neal2018open, jo2018open, sung2019difference}. Within these models, generative networks are designed to produce samples that, when labelled into the $K + 1$'th category, would generalise the open space. However, it is essential to note that OSR models also fix the $K + 1$'th category to the zero'th vector, similar to FM-GANs. Furthermore, OSR-GANs also use regularisation methods to realise the $K + 1$'th category. Our hypothesis states that the generated samples from OSR-GANs are theoretically equivalent to those produced by SSL-GANs (i.e. they all lie in the complementary space). We consider the OSR-designed ARP-GANs by Chen et al.~\cite{chen2021adversarial} since these models hold state-of-the-art OSR-GAN results by significant degrees. Specifically, ARP-GANs constrain the open space within the centre of the embedding space by making use of GANs and reciprocal point learning.

Let each of the $K$ labelled categories be assigned a reciprocal point, $P^j$, which is a prototype (i.e. cluster centre) of everything that category $j$ is not. More specifically, $P^j$ is a learnable feature vector that must be updated until it lies in the centre of the combined embedding space of all other labelled categories ($\neq j$) and the open space. The probability of a sample belonging to category $j$ can then be defined as the distance of the sample to the category's reciprocal point, $P^j$. More specifically, samples are classified into the category with the farthest reciprocal point, as determined by a chosen distance metric, i.e. \mbox{$p_{\text{ARP}}(y = j | x) = \frac{\text{exp}[d(C(x), P^j)]}{\sum_{i=1}^{K} \text{exp}[d(C(x), P^i)]}$}, where \mbox{$d(C(x), P^j)$} is the distance defined by a chosen metric between the reciprocal point and the classifier's output for input sample $x$. To produce such probability scores, classifiers within reciprocal point learning have a different output design than conventional supervised networks.  

Reciprocal point learning does not rely on general neural network classification techniques (i.e. one-hot encoded labels and a cross-entropy loss). Instead, the distance, \mbox{$d(C(x), P^j)$}, between input samples and reciprocal points represent the classifier's logits. Therefore, to ensure uniformity within $C(x)$ and $P^j$, the classifier's output must match the dimensions of the reciprocal points (a vector of size 128 within the original study). The authors then defined the distance metric as the difference between the Euclidean distance and the dot product of the input sample's output and the various reciprocal points - i.e. \mbox{$d(C(x), P^j) = d_e(C(x), P^j) - d_d(C(x), P^j) = (\frac{1}{m} \; \cdot \; || C(x) - P^j ||^2_2) \; \; - \; \; (C(x) \bm{\cdot} P^j)$}. During training, classifiers maximise \mbox{$d(C(x), P^j)$} between labelled training samples and their categories' corresponding reciprocal points. Consequently, reciprocal points and their corresponding categories are pushed to the opposite edges of the classifier's embedding space. However, this learning mechanism does not yet generalise the open space. 

To generalise the open space, an adversarial mechanism is created to bind the Euclidean distance (\mbox{$d_e(C(x), P^j)$}) between labelled training samples and their corresponding reciprocal points to a learned range, $R$. In other words, the classifier maximises $d(C(x), P^j)$ to push the boundaries of labelled categories to the edges of the embedding space, yet, it constrains the distance of these boundaries and their corresponding reciprocal points to remain smaller than $R$. By doing so for multiple labelled categories (i.e. $K > 1$), the open space will be bound to the centre of the embedding space in between all labelled categories (see Fig.~3 in~\cite{chen2021adversarial}). This mechanism is considered adversarial since 1) it pushes the boundaries towards the edges and 2) it pulls them back to the centre to ensure they do not breach $R$. In this manner, adversarial reciprocal point (ARP)-learning is developed. With \mbox{$\gamma = 0.1$}, the adversarial loss function for the classifier network in ARP-learning is described as: 

\begin{equation}
\begin{split}
    \underset{\text{ARP}}{\text{C-Loss}} = - \; \mathbb{E}_{(x, y) \sim D_{\text{lab-train}}}
    & \big[\log(p_{\text{ARP}}(y = y|x)) \\
    & + \gamma \cdot \text{max}(d_e(C(x), P^y) - R, 0)\big].
    \label{eq:cost_c_arp}
\end{split}
\end{equation}

Note that $\text{C-Loss}$ in eq.~\ref{eq:cost_c_arp} is the categorical cross-entropy equivalent for a supervised classifier based on reciprocal points. However, this setup does not yet contain a GAN. To further improve results, an additional GAN model can be appended to the classifier to create a three-player game, i.e. with a classifier, a discriminator and a generator. The discriminator in ARP-GANs is left unchanged to the original GAN (see eq. \ref{eq:cost_d_arpgan} below). However, the ARP-GAN generator is tasked to match a mixture distribution between the real-training data and the reciprocal points, creating a so-called confused generator. Thus, the generator aims to match the real data distribution by learning from the discriminator (as per original GANs) but also pulls generated samples towards the learned reciprocal points. Subsequently, by minimising the distance between all reciprocal points and the generated samples, the classifier can further constrain the open space for improved OSR results. 

To include the additional GAN criterion within ARP-learning, ARP-GANs use the information entropy loss. Specifically, given \mbox{$S(G(z), P^j) = p_{\text{ARP}}(y = j | G(z))$}, the information entropy loss between generated samples and reciprocal points is defined as \mbox{$I(G(z)) = - \sum^{K}_{i = 1} S(G(z), P^i) \cdot \log(S(G(z), P^i))$}. The generator and classifier maximise $I(G(z))$ to pull generated samples towards the reciprocal points and reciprocal points towards generated samples. Considering that the classifier's predictions, \mbox{$p_{\text{ARP}}$}, are dependant on the various reciprocal points, we argue that $I(G(z))$ acts as a regularisation loss function. More specifically, the further a reciprocal point is from samples that belong to its corresponding category, the more confident the classifier is in its predictions. Thus, the classifier's predictions are regularised by pulling reciprocal points towards the generated samples. We also note that regularisation has previously been argued ideal for OSR-GANs~\cite{jo2018open}. Given that the generated samples represent the open-space, this regularisation loss improves the classifier's OSR performance. The three loss functions for the discriminator, generator and classifier in ARP-GANs are described as:

\begin{equation}
\begin{split}
    \underset{\text{ARP-GAN}}{\text{D-Loss}} = 
    & - \; \mathbb{E}_{(x) \sim D_{\text{train}}} [\log(D(x))] \\
    & - \; \mathbb{E}_{z \sim \mathbb{P}^{\text{(z)}}} [\log(1 - D(G(z)))],
    \label{eq:cost_d_arpgan}
\end{split}
\end{equation}

\begin{equation}
    \underset{\text{ARP-GAN}}{\text{G-Loss}} = - \; \mathbb{E}_{z \sim \mathbb{P}^{\text{(z)}}} [\log(D(G(z))) \; + \; I(G(z))],
    \label{eq:cost_g_arpgan}
\end{equation}

\begin{equation}
\label{eq:cost_c_arpgan}
\begin{split}
\underset{\text{ARP-GAN}}{\text{C-Loss}} = 
& - \; \mathbb{E}_{z \sim \mathbb{P}^{\text{(z)}}} [I(G(z))] \\
& \; \begin{split} - \; \mathbb{E}_{(x, y) \sim D_{\text{lab-train}}} &\big[\log(p_{\text{class-ARP}}(y = y|x)) \\
& + \gamma \cdot \text{max}(d_e(C(x), P^y) - R, 0)\big]. \end{split}
\end{split} 
\end{equation}

An ARP-GAN achieves the following: 1) it pushes the embedded classification boundaries of labelled categories towards the edges of the embedding space, 2) it constrains the open space within the centre of the embedding space, 3) it generates samples that represent the open space and 4) it regularises the classifier by pulling reciprocal points towards generated samples. Considering that the generator maximises $I(G(z))$, it is essential to note that the generated samples are also pulled out of the open space and towards the reciprocal points. Therefore, generated samples either lie in the open space or between the open space and the reciprocal points (see Fig. 7 in \cite{chen2021adversarial}). Consequently, we postulate that the generated samples from ARP-GANs also lie in the complementary space similar to those from FM-GANs. Considering that both models also regularise their classifiers with the generated samples, we hypothesis that FM-GANs and ARP-GANs function in the same manner. To prove this hypothesis, we place both models under the same experimental conditions. 

%+- 1000 words
\section{Experiments}
\begin{table*}[t]
\centering
%\begin{tabular}{m{3.2cm} m{2.2cm} m{2.2cm} m{0.3cm} m{2.2cm} m{2.2cm}}
\begin{tabular}{p{3.2cm} p{2.2cm} p{2.2cm} p{0.3cm} p{2.2cm} p{2.2cm}}
\hline
 \multicolumn{1}{c}{} & \multicolumn{2}{c}{\textbf{Semi-supervised}} & \multicolumn{3}{c}{\textbf{Supervised}} \\
\hline
    & SVHN & CIFAR10 & & CIFAR10$\dagger$ & CIFAR10$\star$ \\
Labels per category & 100     & 400  & & Full  & Full \\
\hline
\textbf{Supervised baselines} & & & & &  \\
Softmax & 13.83 $|$ 41.18 & 62.66 $|$ 67.04 &   & 90.24 $|$ 83.79 & 93.36 $|$ 85.26 \\
ARP     & 10.32 $|$ 53.65 &  61.54 $|$ 69.46 &   & 89.19 $|$ 84.53 & 93.90 $|$ 90.10   \\
\hline
\textbf{GANs} & & & & &  \\
FM-GAN  & \textbf{83.40} $|$ 88.02 & \textbf{78.70} $|$ 75.17  &  & \textbf{92.23} $|$ 85.91  & \textbf{95.62} $|$ 87.75   \\
ARP-GAN &  81.51 $|$ \textbf{90.16} &  77.62 $|$ \textbf{77.02} &  & 90.13 $|$ \textbf{87.07} & 94.50 $|$ \textbf{91.00}   \\
&  &   &   &  &   \\
Bad-GAN~\cite{dai2017good}  & \textit{96.75} $|$ - - - &  \textit{86.59} $|$ - - - &   &  &   \\
Triple-GAN~\cite{li2021triple}  & \textit{96.55} $|$ - - - &  \textit{89.99} $|$ - - - &   &  &   \\
Margin-GAN~\cite{NEURIPS2019_517f24c0} & \textit{94.86} $|$ - - - &  \textit{93.56} $|$ - - -  & & &  \\
Negative-GAN~\cite{jo2018open} &  &   &  & - - - $|$ \textit{72.90}  &     \\
Counter-GAN~\cite{neal2018open} &  &  &   &  &  - - -  $|$ \textit{83.80} \\
\hline
\end{tabular} \\ 
\begin{tablenotes}
%\centering
\item \textit{\doublespacing Results are averaged over five randomized trials}
\end{tablenotes}
\caption{Results for the SSL-OSR experiments in part one. The number of labelled training samples per labelled category are indicated. Each result is split as $a \; | \; b $ with $a$ representing the percentage accuracy for labelled categories and $b$ representing the $\text{AUROC} * 100$. The top-performers are shown in bold. For ease of reference, the bottom five GANs are previously published results using their own code-bases (i.e. different optimisation schemes and hyperparameter choices).}
\label{table:1}
\end{table*}

Our experiments are divided into two parts. In the first part, we compare FM-GANs and ARP-GANs under the same experimental conditions. These results will prove our hypothesis that SSL-GANs and OSR-GANs operate similarly. Given this conclusion, part two places several other GAN models under SSL-OSR experiments. These experiments are intended to create benchmark results and provide further knowledge on applying GANs in the SSL-OSR setting. 

\subsection{Part One: FM-GANs and ARP-GANs}
To prove our hypothesis that FM-GANs and ARP-GANs operate in the same manner, we must show that these models achieve similar results for identical experimental setups. Both GAN models must be placed under the same experimental settings to ensure good scientific practice. Specifically, we opt to integrate FM-GANs into the ARP-GAN code base to ensure the same optimisation techniques and hyperparameters are used for both models. However, it must be noted that FM-GANs only use two networks, while ARP-GANs use three. Thus, ARP-GANs are considered more computationally complex. For evaluation, we indicate the accuracy scores over the labelled categories and the area under the receiver operating characteristic (AUROC) curve for unobserved novelty detection~\cite{geng2020recent}. As per OSR standards, the prediction scores of samples from labelled categories are used to gather a range of threshold values to create the ROC. Subsequently, the AUROC provides a numerical metric of the resulting graph. 

All experiments use the SVHN or CIFAR10 datasets, with training and testing setups per conventional SSL and OSR protocols. For the SSL settings, 100 and 400 labelled training samples are provided per category for SVHN and CIFAR10, respectively, with the remaining samples used as unlabelled training samples. During testing, the CIFAR10 test set is appended to the SVHN test set to introduce unobserved novel categories within the SVHN experiments, and the CIFAR100 test set is appended to the CIFAR10 test set to introduce unobserved novel categories within the CIFAR10 experiments. For the fully-supervised settings (i.e. all training labels are revealed as is conventional for OSR), we conduct two experiments using the CIFAR10 dataset. The CIFAR10$\dagger$ experiment uses all training samples from the CIFAR10 dataset while testing samples are from CIFAR10 and CIFAR100. Then, for the CIFAR10$\star$ experiment, $K = 6$ random categories from CIFAR10 are defined as labelled categories, while the remaining $4$ categories are used as unobserved novel categories. 

The FM-GAN and ARP-GAN results are shown in Table \ref{table:1}. We also indicate the performance of two supervised baseline models, traditional softmax classifiers and ARP classifiers, noting that these models can only access the labelled training samples. These baselines were also trained using the ARP-GAN code base for equal optimisation and hyperparameter choices. Immediately, we note that the FM-GAN and ARP-GAN improve over the baselines for all experiments. Thus, we can conclude that adding the GAN to a classifier's training scheme improves the model's accuracy. Furthermore, we can see that FM-GANs (as expected) and ARP-GANs drastically improve over the supervised baselines under the SSL settings. Specifically, the supervised baselines do not use the unlabelled training samples, unlike FM-GANs and ARP-GANs. Thus, the increase in performance for the SSL experiments over the baselines concludes that FM-GANs and ARP-GANs are both SSL models. 

It is also interesting to discuss the SSL SVHN experiments wherein the supervised baselines fail to learn a representation of the domain. SVHN is a dataset wherein each category is defined as a street house digit from 0 - 9. However, considering that the images are cropped versions of larger images that contain multiple digits, SVHN data samples regularly contain multiple digits within a single image. Consequently, the label of an image corresponds to the digit in the centre of the image, yet the image might contain multiple other digits surrounding it. Therefore, when only 100 labelled training samples are provided to the classifier (with no unlabelled samples), the classifier will fail to distinguish between digits in the centre of the image and other digits around it. However, when the classifier also has access to unlabelled training samples (i.e. for FM-GANs and ARP-GANs), then the model can accurately learn a representation of the domain. This result clearly indicates the benefit of SSL over supervised learning, beyond even the cost-benefit. 

Finally, when comparing FM-GANs and ARP-GANs across all experiments, it is clear that both models achieve near-identical results for both SSL and OSR. However, we note that FM-GANs produce slightly higher SSL results, and ARP-GANs produce slightly higher OSR results. These results are expected since FM-GANs were explicitly designed for SSL and ARP-GANs for OSR. Nevertheless, we find the near-identical results as satisfactory evidence to conclude that FM-GANs and ARP-GANs operate similarly. However, we note that the SSL results for these models do not compare well against the current state-of-the-art SSL-GANs. Specifically, because we used the ARP-GAN code base, our models applied optimisation techniques for OSR and not SSL. However, given our conclusion of the SSL-OSR link, we are now warranted to apply state-of-the-art SSL-GANs on the combined SSL-OSR task. 

\subsection{Part Two: Other GANs}
Considering the proven link between SSL-OSR, experiments using state-of-the-art SSL-GANs on SSL-OSR would provide reasonable benchmark results. These results will also provide further insights into the inner workings of SSL-OSR GANs. We experiment with three different SSL-GANs: Bad-GANs, which cemented the theory of the complementary space~\cite{dai2017good}; Margin-GANs, which applied a new theory of margins within SSL-GANs~\cite{NEURIPS2019_517f24c0}; and Triple-GANs, which present a categorical approach to SSL~\cite{li2021triple}. These experiments are not meant to compare the different GAN models. Instead, we use each study's publicly available code base, operating under its own optimisation and hyper-parameter choices. Therefore, the results are only for comparisons in future studies. Furthermore, by noting the differences between each model, we develop a greater understanding of SSL-OSR using GANs. 

It is important to discuss several differences between each GAN model. First, on computational complexity, Bad-GANs deploy a two-player game similar to FM-GANs, while Margin-GANs and Triple-GANs deploy a three-player game similar to ARP-GANs. Furthermore, Bad-GANs use considerably smaller networks than either Margin-GANs and Triple-GANs, noting that Margin-GANs opt for a more complicated classifier network than Triple-GANs, while Triple-GANs opt for more complicated discriminator and generator networks than Margin-GANs. Thus, Bad-GANs are the least computationally complex, with Margin-GANs second and Triple-GANs the most computationally complex. Beyond architecture, it is also important to note that Margin-GANs and Triple-GANs enhance their proposed models by deploying the Mean-Teacher SSL model~(\cite{tarvainen2017mean}) in conjunction with their GANs. Considering that Bad-GANs do not make use of Mean-Teacher models, Margin-GANs and Triple-GANs can be expected to perform better, noting the additional complexity of deploying Mean-Teachers.  

We conduct several SSL-OSR experiments. First, the same SVHN and CIFAR10 experiments are run, as per part one. Then, we run two more difficult experiments using the CIFAR100 dataset. It is important to note that CIFAR100 only has $500$ training samples per category (compared to $4000$ for CIFAR10) and has $100$ categories within its domain (compared to $10$ for CIFAR10). Two experiments are designed as follows: 1) the CIFAR100$\dagger$ experiment uses $20$ categories as labelled categories and the remaining $80$ as unobserved novel categories, and 2) the CIFAR100$\star$ uses $50$ categories as labelled categories and the remaining $50$ as unobserved novel categories. For both experiments, we only provide $100$ labelled training samples for each labelled category, leaving the remaining $400$ samples as unlabelled. The results for the SVHN, CIFAR10 and CIFAR100 experiments are shown in Table~\ref{table:2}. 

It is immediately interesting to note that the Bad-GAN and Triple-GAN SSL results from our SVHN and CIFAR10 experiments are slightly lower than their published results (see Table~\ref{table:1}). However, concerning our model setups, we kept the publicly available code from each GAN's respective studies unchanged. More concretely, we could not recreate their published results, which places these studies' publicly available code into question. Nevertheless, the comparisons between the various GAN models still indicate unique insights into the SSL-OSR link. Immediately, we can conclude that Margin-GANs hold the new state-of-the-art results for SSL-OSR. Although Triple-GANs slightly outperformed Margin-GANs in the SVHN experiments, the remaining results indicate the Margin-GANs are superior in performance over the other GAN models. A closer look at each model provides insight into their performance. 

Margin-GANs use the most computationally complex classifying network of all three GAN models. However, considering that Triple-GANs use the most computationally complex GAN, our results indicate that a smaller GAN but larger classifying network is more beneficial to SSL-OSR results. Interestingly, Bad-GANs perform the worst in all experiments. Several factors might contribute to these results: 1) Bad-GANs rely on the discriminator acting as a classifier, 2) Bad-GANs do not make use of the mean-teacher SSL model, and 3) Bad-GANs make use of the least computationally complex classifying network. Future studies on Bad-GANs should consider a three-player game similar to Margin and Triple-GANs, including the mean-teacher model within its setup and increasing the classifying network size of the model. In other words, a unified approach of the various GAN models is worth further exploring. 

It is also worth noting the decreasing performance between the various experiments. Specifically, our results clearly show that the CIFAR100 experiments are more complex than the SVHN and CIFAR10 experiments. The lower performance for the CIFAR100 experiments can be attributed to 1) fewer training samples per category and 2) a larger number of categories (i.e. $K$). Regarding the smaller number of training samples, it is interesting that Margin-GANs outperform Triple-GANs in these settings since Triple-GANs were argued to be useful in low-data regimes. Nevertheless, we again see the larger classifying network of Margin-GANs achieving top results while decreasing the least in performance between the two CIFAR100 experiments. Finally, we note the comparisons between these GAN models and the OSR-optimised ARP-GANs of Part One above. We can see that these SSL-optimised GANs far-outperform ARP-GANs at both SSL and OSR. 

\begin{table*}[t]
\centering
%\begin{tabular}{m{3.2cm} m{2.2cm} m{2.2cm} m{0.3cm} m{2.2cm} m{2.2cm}}
\begin{tabular}{ p{3.5cm} p{2.2cm} p{2.2cm} p{2.2cm} p{2.2cm}}
\hline
 & \multicolumn{3}{c}{\textbf{Semi-supervised}} \\
\hline
    & SVHN & CIFAR10 & CIFAR100$\dagger$ & CIFAR100$\star$ \\
Labelled categories ($K$) & 10 & 10  & 20 & 50 \\
Labels per category & 100 & 400  & 100 & 100 \\
\hline
\textbf{GANs} & & &  \\
Bad-GAN & 92.29 $|$ 62.96 &  80.52 $|$ 60.17 &  61.09  $|$ 63.22 &  51.52  $|$ 58.24  \\
Margin-GAN & 94.57 $|$ 97.73 &  \textbf{92.93} $|$ \textbf{88.91} & \textbf{79.85}  $|$ \textbf{78.39} &  \textbf{75.76}  $|$ \textbf{75.18} \\
Triple-GAN &  \textbf{94.86} $|$ \textbf{98.18} &  89.04 $|$ 86.34 & 74.17  $|$ 72.46  &  64.58  $|$ 68.99  \\
\hline
\end{tabular} \\
\begin{tablenotes}
%\centering
\item \textit{\doublespacing Results are averaged over three randomized trials}
\end{tablenotes}
\caption{Results for the SSL-OSR experiments in part two. The number of labelled categories and the number of labelled training samples per labelled category are indicated. Each result is split as $a \; | \; b $ with $a$ representing the percentage accuracy for labelled categories and $b$ representing the $\text{AUROC} * 100$. The top-performers are shown in bold.}
\label{table:2}
\end{table*}

\begin{figure*}[!t]
\centering
\subfloat[Real images]{\includegraphics[width=0.45\textwidth]{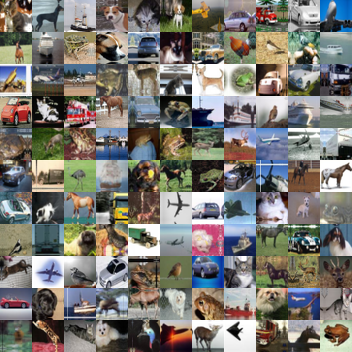}
\label{fig:5a}}
\hfil
\centering
\subfloat[Generated images]{\includegraphics[width=0.45\textwidth]{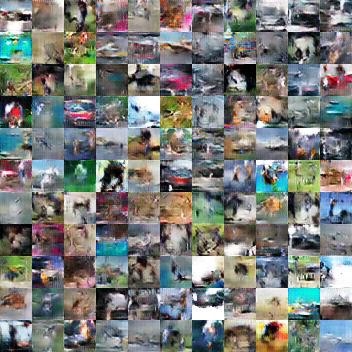}
\label{fig:5b}}
\caption{Comparing real and generated images from SSL Margin-GANs - (a) shows examples of real CIFAR10 images; (b) shows examples of generated images from the trained Margin-GAN on the SSL CIFAR10 experiment. }
\label{fig:5}
\end{figure*}

To conclude our experiments, it is interesting to view the generated samples produced by a trained Margin-GAN. In Fig.~\ref{fig:5}, we present examples of real CIFAR10 training images and generated images from the Margin-GAN CIFAR10 experiment. It is clear that the images produced are of visibly lower quality compared to the real images. Formally, the generated images have indistinct features, lack sharp definition and have a blurred and amorphous appearance. This demonstrates that the generated images do not conform to the real training distribution, which matches the notion of generating bad-looking samples. However, these samples do still produce similar qualities (e.g. colors and contrast) to the real images, as would be required for the images to lie in the complementary space. It is clear that regularising classifying networks with samples that lie in the complementary space allows the classifier network to achieve both SSL and OSR. 

%+- 850
\section{Discussion}
Our results clearly demonstrate the intrinsic connection between SSL and OSR GAN based methods. Classifiers capable of simultaneously detecting unobserved novel categories can now be trained using significantly fewer labeled training samples. Novelty detection in and of itself is very useful within a broad number of applications. For example, contamination, leaks or faults cause unusual patterns in fluid dynamics~\cite{arqub2023adaptive, engelbrecht2019modeling}, and novel diseases can appear over time within any medical domain~\cite{maayah2022numerical, maayah2022multistep}. In fact, considering that all natural phenomena act over the temporal dimension, novel categories can appear in any inductive setting~\cite{badawi2023stochastic, engelbrecht2020open}. With our results, detecting and separating novel categories can now be conducted using a far more cost-efficient approach. 

Novelty detection also stands to benefit the general data cleaning process, which is critical for the optimisation of statistical modelling. Consider any general statistical model, e.g. linear regression or k-mean clustering, which are common in data science. These models are easily skewed by anomalies and novel categories. Thus, practitioners should either conduct a similar extensions of these models to OSR, as done in this study, or practitioners should use our proposed model as a pre-processing data-cleaning step. In turn, modelling would improve in accuracy and reduce the cost in domain specialist domains - e.g. material sciences~\cite{almessiere2022tuning, almessiere2022investigation, gibbons2023metal}, transport and logistics~\cite{pevceny2020optimisation}, and grid optimisation~\cite{zhang2019deep}. However, future SSL-OSR research should also take several other factors into consideration. 

Generally, SSL and OSR research consider balanced data settings. However, in real-world scenarios, it is uncommon for datasets to contain the same number of training samples for every category. Future studies should consider bridging SSL-OSR to label-imbalance settings where the number of labels is heavily skewed towards certain categories. However, GANs also hold promise for such settings since a GAN is able to generate realistic looking data (noting the contrast to this study) to balance the number of labels~\cite{hwang2019hexagan}. Another factor that future studies might consider are multi-modal distributions. In the context of classification, multi-modal distributions are described as categories that prescribe to multiple output labels~\cite{wang2021can}. Multi-modal datasets regularly occur in real-world environments, and so SSL-OSR should also consider these cases. 

It is also important for future research to consider another type of novel category that exists in partially labelled training sets. More specifically, observed novel categories are in contrast to unobserved novel categories since these appear scattered in unlabelled training sets~\cite{engelbrecht2020learning, da2014learning}. Although this study focuses on conventional SSL (i.e. without observed novel categories), extending the combined framework of SSL-OSR to also consider observed novel categories would further enhance the practicality and cost-efficiency of classifier development~\cite{engelbrecht2020open}. By detecting multiple novel category types, practitioners could discover new and interesting patterns under a combined framework of class-incremental learning and co-operative artificial intelligence~\cite{van2022three, dafoe2021cooperative}.  

\section{Conclusion}
In this study, we proved that semi-supervised learning (SSL) and open-set recognition (OSR) models based on generative adversarial networks (GANs) are linked by the same theory applied to their classification networks. Although SSL-GANs and OSR-GANs are generally kept separate in literature, we found that both model types require their generators to produce samples within the complementary space, which are then used to regularise their classifiers. Subsequently, SSL-GANs and OSR-GANs generalise the open space within their respective $K + 1$'th categories. Our findings suggest  OSR-GANs are capable at SSL given further optimisation and that SSL-GANs, without modifications, are highly capable at OSR. Furthermore, we find that Margin-GANs produce state-of-the-art results for the combined task of SSL-OSR, noting that this study is the first to set benchmark results for this unified field. Future studies should extend the SSL-OSR framework to consider other real-world factors and apply the SSL-OSR GAN models to various applications. 

\clearpage
\section*{Declaration of Generative AI and AI-assisted technologies in the writing process}
During the preparation of this work the author(s) used Grammarly and ChatGPT in order for grammatical improvement for better readability of the study. After using this tool/service, the author(s) reviewed and edited the content as needed and take(s) full responsibility for the content of the publication.

\printbibliography
  
\end{document}